\documentclass{article} 
\usepackage{iclr2020_conference,times}

\usepackage{amsmath,amsfonts,bm}

\def\eqref#1{equation~\ref{#1}}

\def\1{\bm{1}}

\DeclareMathAlphabet{\mathsfit}{\encodingdefault}{\sfdefault}{m}{sl}
\SetMathAlphabet{\mathsfit}{bold}{\encodingdefault}{\sfdefault}{bx}{n}

\usepackage{hyperref}
\usepackage{url}
\usepackage{graphicx}
\usepackage{multirow}
\usepackage{subcaption}
\usepackage{dblfloatfix}
\usepackage{booktabs}
\usepackage{colortbl}
\definecolor{mygray}{gray}{.9}
\usepackage{cleveref}
\crefname{section}{§}{§§}
\Crefname{section}{§}{§§}

\title{Cross-Supervised Object Detection}

\author{
Zitian Chen\(^1\)
\hspace{.9cm}
Zhiqiang Shen\(^2\)
\hspace{.9cm}
Jiahui Yu\(^3\)
\hspace{.9cm}
Erik Learned-Miller\(^1\)
\\
\(^1\) UMass Amherst
 \hspace{.2cm}
\(^2\) Carnegie Mellon University 
 \hspace{.2cm}
\(^3\)University of Illinois at Urbana-Champaign
}

\newcommand{\secvspace}{\vspace{-0.2em}}
\newcommand{\subsecvspace}{\vspace{-0.2em}}

\iclrfinalcopy 
\begin{document}

\maketitle

\begin{abstract}
After learning a new object category from image-level annotations (with no object bounding boxes), humans are remarkably good at precisely localizing those objects. {However,} building good object localizers (i.e., {\em detectors}) currently requires expensive instance-level annotations. 
While some work {has} been done on learning detectors from weakly labeled samples (with only class labels), these detectors do poorly at localization.
In this work, we show how to build better object detectors from weakly labeled images of new categories by leveraging knowledge learned from fully labeled base categories.
We call this novel learning paradigm {\bf cross-supervised object detection}. We propose a unified framework that combines a detection head trained from instance-level annotations and a recognition head learned from image-level annotations, together with a spatial correlation module that bridges the gap between detection and recognition. These contributions enable us to better detect novel objects with image-level annotations in complex multi-object scenes such as the COCO dataset.\\
  
\end{abstract}

\secvspace
\section{Introduction}
\secvspace

Deep architectures have achieved great success in many computer vision tasks including object recognition and the closely related problem of object detection.
Modern detectors, such as the Faster RCNN (\cite{faster_rcnn}), YOLO (\cite{YOLO}), and RetinaNet (\cite{focalLoss}), use the same network backbone as popular recognition models.
However, even with the same backbone architectures, detection and recognition models require different types of supervision. A good detector relies heavily on precise bounding boxes and labels for each instance (we shall refer to these as  \textit{instance-level annotations}) , whereas a recognition model needs only image-level labels.
Needless to say, it is more time consuming and expensive to obtain high quality bounding box annotations than class labels. As a result, current detectors are limited to a small set of categories relative to their object recognition counterparts. To address this limitation, it is natural to ask, `` Is it possible to learn  detectors with only class labels?'' This problem is commonly referred to as weakly supervised object detection (WSOD).


Early WSOD work (\cite{hoffman2014lsda}) showed fair performance by directly applying recognition networks to object detection. More recently, researchers have used multiple instance learning methods (\cite{dietterich1997solving}) to recast WSOD as a multi-label classification problem (\cite{bilen2016wsddn}). However, these weakly supervised detector perform poorly at localization. Most WSOD experiments have been conducted on the ILSVR (\cite{ILSVRC15}) data set, in which images have only a single object, or on the PASCAL VOC (\cite{everingham2010pascal}) data set, which has only 20 categories. The simplicity of these data sets limits the number and types of distractors in an image, making localization substantially easier. Learning from only class labels, it is challenging to detect objects at different scales in an image that contains many distractors. In particular, as shown in our experiments, weakly supervised object detectors do not 
work well in \emph{complex} multi-object scenes, such as the COCO dataset (\cite{COCO}). 

To address this challenge, we propose a new form of learning in which the localization of classes with only object labels (weakly labeled classes) can benefit from other classes that have ground truth bounding boxes (fully labeled classes). We refer to this new learning paradigm as {\em cross-supervised object detection} (CSOD).
More formally, we define CSOD as follows. At training time, we are given 1) images contain objects from both base and novel classes, 2) both class labels and ground truth bounding boxes for base objects, and 3) only class labels for novel objects.
Our goal is to detect novel objects.
In CSOD, base classes and novel classes are disjoint. Thus, it can be seen as performing fully-supervised detection on the base classes and weakly supervised detection on the novel classes. 
It has similarities to both transfer learning and semi-supervised learning, since it transfer knowledge from base class to novel class and have more information about some instances than other instances.
However, CSOD represents a distinct and novel paradigm for learning.

\begin{figure*}
  \centering
  \includegraphics[width=12cm]{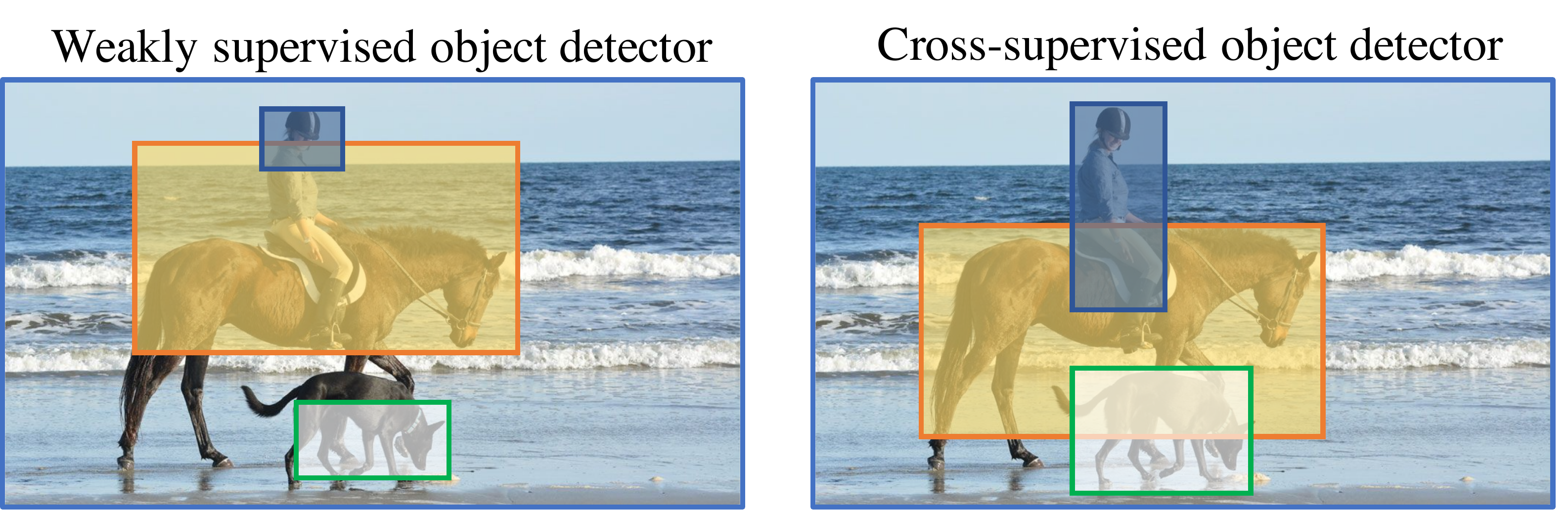}
  \caption{\textbf{A comparison between weakly supervised object detector and our detector.} Weakly supervised object detector only detects the most discriminating part of an object, e.g., focus on head of a person when detecting a person; or being distracted by co-occurring instances, e.g., distracted by the person on the horse when detecting a horse. Our detector can address these issues. }
  \label{fig:drawback}
\end{figure*}

The  current  state of affairs for this problem is directly applying weakly supervised object detection methods.
However, these approach ignores knowledge about localization learned from base classes and has several drawbacks. As shown in Fig.~\ref{fig:drawback}, a weakly supervised object detector tends to detect only the most discriminating part of  novel objects instead of the whole object. Notice how only the head of the person, and not the whole body, is detected. Another issue is that the localizer for one object (e.g., the horse) may be confused by the occurrence of another object, such as the person on the horse. 
This example illustrates the gap between detection and recognition: without ground truth bounding boxes, the detector acts like a standard recognition model -- focusing on discriminating rather than detecting.

In this paper, we explore two major mechanisms for improving on this. Our first mechanism is unifying detection and recognition. Using the same network backbone architecture, recognition and detection can be seen as image-level classification and region-level classification respectively, suggesting a strong relation between them. In  particular, it suggests a shared training framework in  which  the same backbone is used with  different heads for detection and recognition. Thus, we combine a detection head learned from ground truth bounding boxes, and a recognition head learned in a weakly supervised fashion from class labels. Unlike a traditional recognition head, our recognition head produces a class score for multiple proposals and is capable of detecting objects. 
The second mechanism is learning a \textit{spatial correlation module} to reduce the gap between detection and recognition. {It takes several high-confidence bounding boxes produced by the recognition head as input, and learns to regress ground truth bounding boxes.} By combining these mechanisms together, our model outperforms all previous models when all novel objects are weakly labeled.

In summary, our contributions are three-fold. First, we define a new task---cross-supervised object detection, which enables us to leverage knowledge from fully labeled base categories to help learn a robust detector from novel object class labels only. Second, we propose a unified framework in which two heads are learned from class labels and detection labels respectively, along with a spatial correlation module bridging the gap between recognition and detection. Third, we significantly outperform existing methods (\cite{zhang2018mixed,oicr,tang2018pcl}) on PASCAL VOC and COCO, suggesting that CSOD could be a promising approach for expanding object detection to a much larger number of categories. 

\secvspace
\section{Related Work}
\secvspace

\textbf{Weakly supervised object detection.} 
WSOD (\cite{kosugi2019object,zeng2019wsod2,yang2019towards,CMIL,Arun_2019_CVPR,wan2018min,zhang2018zigzag,zhu2017soft,zhang2018w2f,li2019weakly,gao2019c,kosugi2019object}) attempts to learn a detector with only image category labels. Most of these methods adopt the idea of Multiple Instance Learning (\cite{dietterich1997solving}) to recast WSOD as a multi-label classification task. \cite{bilen2016wsddn} propose an end-to-end network {by modifying a classifier} to operate at the level of image regions, serving as a region selector and a classifier simultaneously. \cite{oicr} and \cite{tang2018pcl} find that {several iterations of online refinement} based on the outputs of previous {iterations} {boosts} performance. \cite{wei2018ts2c} and \cite{diba2017weakly} use semantic segmentation based on class activation maps (\cite{CAM}) to help generate tight bounding boxes. 
However, WSOD methods tend to focus on the most discriminating part of {an} object and {are prone to distractions} from co-occurring objects. {Detecting a} part of the object or distractors represents {convergence} to a local optimum. Thus, their performance depends heavily on initialization.
In comparison, our proposed cross-supervised object detector {alleviates the issue of getting trapped in a local optimum} by leveraging knowledge learned from fully labeled base categories.

\textbf{Cross-supervised object detection.} 
There are several previous works using both image-level and instance-level annotations.
\cite{kuen2019scaling} learned a parameter transferring {function} between a classifier and a detector, enabling an image-based classification network to be adapted to a region-based classification network. \cite{hoffman2014lsda} and \cite{tang2016large} {propose methods of adaptation for knowledge transfer from classification features to detection features}. \cite{uijlings2018revisiting} use a proposal generator trained on base classes to transfer knowledge by leveraging a MIL framework, organized in a semantic hierarchy. \cite{hoffman2015detector} design a three-step framework {to} learn a feature representation from weakly supervised classes and strongly supervised classes jointly. However, these methods can only perform object localization in single object scenes such as ILSVRC, whereas our method can perform object detection in complex multi-object scenes as well, e.g. COCO. 
\cite{gao2019note} use a few instance-level labels and a large scale of image-level labels for each category in a training-mining framework, which is referred to as semi-supervised detection. \cite{zhang2018mixed} propose a framework named MSD that learn objectness on base categories and use it to reject distractors when learning novel objects. In comparison, our spatial correlation module not only learns objectness, but also refines coarse bounding boxes. Further, our model learns from both base and novel classes instead of only novel classes.
\begin{figure*}
  \centering
  \includegraphics[width=12cm]{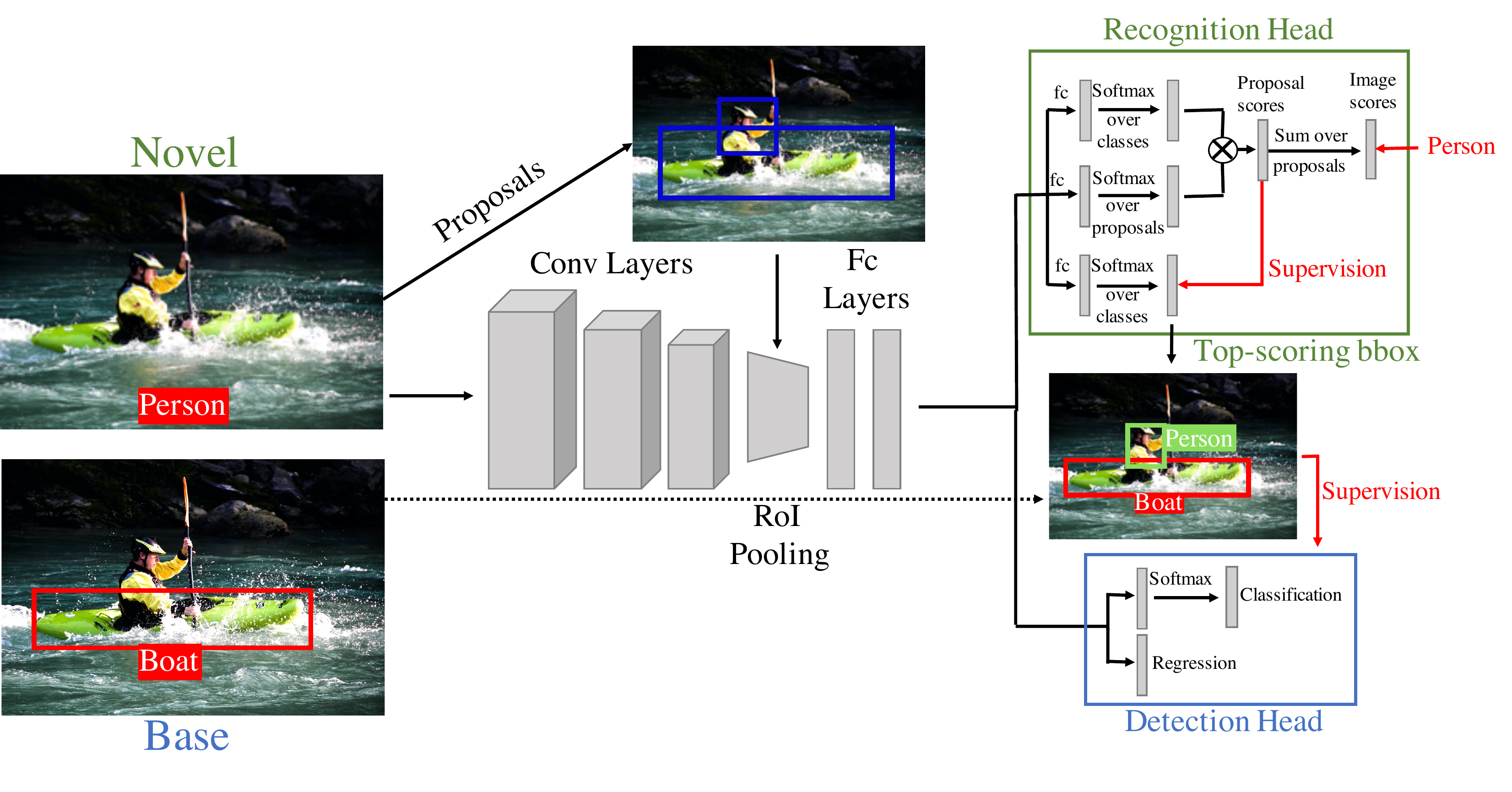}
  \caption{\textbf{Our Detection-Recognition Network (DRN) without the spatial correlation module.} In this {illustration}, \textit{Person} belongs to novel classes and \textit{Boat} belongs to base classes. {The} recognition head {learns} from {the} class label {\textit{Person}} and {outputs the} top-scoring bounding box to help {the} detection head learn to detect {the} person. {The spatial correlation module, discussed in \cref{SCM}, can be added to further refine the top-scoring bounding boxes.}
  }
  \label{fig:DRN}
  \vspace{-0.2cm}
\end{figure*}

\secvspace
\section{Cross-Supervised Object Detection}
\secvspace
\label{split}

CSOD {requires} us to learn from instance-level annotations (detection labels) and image-level annotations (recognition labels). 
In this section, we explain the unification of detection and recognition and introduce our framework. 
In the next section, we describe our novel spatial correlation module.

\subsecvspace
\subsection{Unifying Detection and Recognition} 
\subsecvspace
\label{baseline}
 

\textbf{How to learn a detector from both instance-level and image-level annotations?} 
Since detection and recognition can be seen as region-level and image-level classification respectively, a natural choice is to design a unified framework that combines a detection head and a recognition head {that can learn} from image-level and instance-level annotations respectively. 
Here we exploit several \textit{baselines} to unify the detection and recognition head.
(1) \emph{Finetune}. We first learn through {the} detection head on base classes with fully labeled samples. Then{,} we finetune our model using {the} recognition head {on novel classes} with only class labels. 
(2) \emph{Two Head.} We simultaneously learn {the} detection and recognition head on base and novel classes, respectively. The weights of the backbones are updated using {the} loss {backpropagated} from both heads jointly.
(3) \emph{Two head $^+$.} Instead of {learning only} on novel classes, we learn {the} recognition head {from class labels of both base and novel classes whereas the recognition head remain the same.} 
(4) \emph{Two Branch.} 
Instead of having two shared fully-connected layers after RoI pooling layer (see Fig.~\ref{fig:DRN}), we make these two fully-connected layers seperated, allowing the detection and recognition head to have separate unshared pair of fully-connected layers each.
Everything else is the same as the \emph{Two Head} baseline.
 Experiments are conducted to compare these baselines in \cref{pascal} and \cref{coco}. Our proposed model is based on \emph{Two Head}. We will discuss the details in \cref{DRN}. 

\textbf{The connection between the recognition and detection head.} {The} baselines mentioned above only use {the} recognition head to detect novel objects, ignoring the fact that a detection head can play the same role even better. A majority of WSOD methods (\cite{oicr,CMIL,wei2018ts2c}) find that re-train a new detector taking the top-scoring bounding boxes from a weakly supervised object detector as ground truth marginally improve the performance.
Even with coarse and noisy pseudo bounding boxes, a standard object detector {produces} better detection results than a weakly supervised object detector. Keeping this hypothesis in mind, we introduce a guidance from the recognition head to the detection head. For each of the novel categories existing in a training sample, the recognition head outputs the top-scoring bounding box, which are then used by the detection head as supervision in that sample. 

\subsecvspace
\subsection{Detection-Recognition Network}
\subsecvspace
\label{DRN}

The structure of our Detection-Recognition Network (DRN) is shown in Fig.~\ref{fig:DRN}. Given an image, we first generate 2000 object proposals by Selective Search (\cite{uijlings2013selective}) or RPN (\cite{faster_rcnn}) trained on base classes. The image and proposals are fed into several convolutional (conv) layers {followed by} a region-of-interest (RoI) pooling layer (\cite{girshick2015fast}) to output fixed-size feature maps. Then, these feature maps are fed into two fully connected (fc) layers to produce a collection of proposal {features}, which are further branched into {the} recognition and detection head.
 
\textbf{Recognition Head.} We followed previous WSOD methods to design our recognition head. Since OICR (\cite{oicr}) is simple, neat, and commonly being used, we make our recognition head the same as OICR, but {with fewer refinement branches to reduce the computation cost}. However, our recognition head can be replaced by any WSOD structure as shown in \cref{ablation}. 

Within the recognition head as shown in Fig.~\ref{fig:DRN}, the proposal features are branched into three streams producing three matrices $\mathbf{x}^c,\mathbf{x}^d,\mathbf{x}^e \in \mathbb{R}^{C\times |R|}$, where $C$ is the number of novel classes and $|R|$ is the number of proposals.
Then the two matrices $\mathbf{x}^c$ and $\mathbf{x}^d$ are passed through a softmax function over classes and proposals respectively: $\sigma(\mathbf{x}^c)$
and $\sigma(\mathbf{x}^d)$.
A proposal score $\mathbf{x}^R_{cr}$, indicating the score of $c^{th}$ novel class for $r^{th}$ proposal, corresponds to the respective element of the matrix $\mathbf{x}^R = \sigma(\mathbf{x}^c)\odot \sigma(\mathbf{x}^d)$, where $\odot$ refers to an element-wise product.
Finally, we obtain the image score of $c^{th}$ class $\phi_c$ by summing over all proposals: $\phi_c=\sum_{r=1}^{|R|} x_{cr}^R$. Then we calculate a standard multi-class cross-entropy loss as shown in the first term of Eq.\ref{eqn:rec}. 
Another matrix $\mathbf{x}^e$ is passed through a softmax function over classes, the result of which is expresses as a weighted multi-class cross entropy loss as shown in the second term of Eq.\ref{eqn:rec}.
We set the pseudo label for each proposal $r$ based on its IoU (or overlap) with the top-scoring  proposal of $c^{th}$ class, $y_{cr} = 1$ if IoU $> 0.5$ and $y_{cr} = 0$ otherwise. The weight $w_r$ for each proposal $r$ is its IoU with the top-scoring proposal. The total loss for the recognition head is 
\begin{equation}
L_{rec} =  [-\sum_{c=1}^C {y_c log \phi_c + (1-y_c)log(1-\phi_c) }] + [-\frac{1}{|R|}\sum_{r=1}^{|R|} \sum_{c=1}^{C+1} w_r y_{cr} log x_{cr}^e ] \label{eqn:rec}
\end{equation}

\textbf{Supervision from our recognition head.} We use the {matrix} $x^e$ to {propose} pseudos bounding boxes to guide the detection head. Specifically, we select one top-scoring proposal for each object category that {appears} in the image as a pseudo bounding box, {as done in OICR}. {We introduce the spatial correlation module in \cref{SCM}, to further refine this pseudo ground truth.}

\textbf{Detection Head.} {Now that we have pseudo bounding boxes for novel objects} and ground truth bounding boxes {for} base objects, we train our detection head like a standard detector. For simplicity and efficiency, our detection head use the same structure of Faster R-CNN (\cite{faster_rcnn}). At inference time, the detection head {produces} detection results for both base categories and novel categories. 

\begin{figure*}
  \centering
  \includegraphics[width=13.5cm]{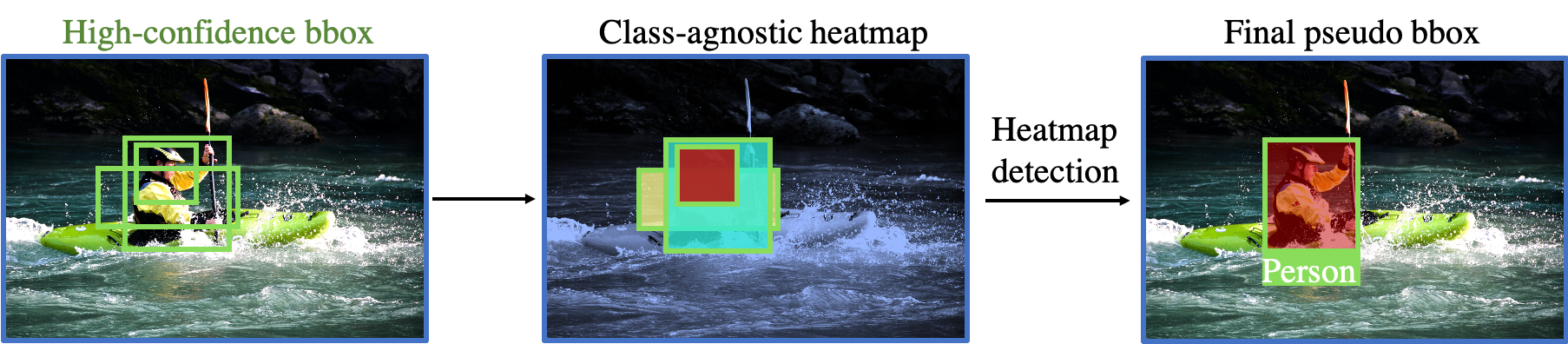}
  \caption{\textbf{Our spatial correlation module (SCM).} Our SCM learns to capture spatial correlation {among} high-confidence bounding boxes, {generating a class-agnostic heatmap for the whole image. A heatmap detector is then trained to learn ground truth bounding boxes.}}
  \label{fig:transfer}
  \vspace{-0.2cm}
\end{figure*}

\secvspace
\section{Learning to Model Spatial Correlation}
\secvspace
\label{SCM}

Our intuition is that there exists spatial correlation {among} high-confidence bounding boxes{,} and such spatial correlation can be captured to predict ground truth bounding boxes. By representing the spatial correlation in a class-agnostic heatmap, we can easily learn a mapping from recognition-based bounding boxes to ground truth bounding boxes {for base categories, and then transfer this mapping to novel categories}. 

Thus, we propose a spatial correlation module (SCM). {SCM is used as a guidance refinement technique, taking sets of high-confidence bounding boxes from the recognition head, and correspondingly returning pseudo ground truth bounding boxes to the detection head. These pseudo ground truth boxes act as supervision while training on novel categories.} The framework of SCM is showed in Fig.~\ref{fig:transfer}. {Within this module, we} first generate a class agnostic heatmap based on the high-confidence bounding boxes predicted by our recognition head, and then we perform detection on top of the heatmap.

\textbf{{Heatmap synthesis.}} {We want to capture the information about how the high-confidence bounding boxes interact amongst themselves. Here, we introduce a simple way of achieving this using a class-agnostic heatmap.} For each category  existing in the image {$y_c=1, c \in C$}, we first threshold and select high-confidence bounding boxes of class $c$. Then we synthesize a corresponding class-agnostic heatmap, which is {essentially} a two-channel feature map {of} the same size as the original image. The value at each pixel is the sum and the maximum of confidence over all selected bounding boxes covering that pixel.

\textbf{Heatmap detection.} {We consider each class-agnostic heatmap as a two-channel image, and perform detection on it}. Specifically, we learn a class-agnostic detector on base classes, {that we} further {use} to produce pseudo {ground truth} bounding boxes for novel objects. 

{For this task,} we use a lightweight one-stage detector, {consisting} of only five convolutional layers. We follow the same network architecture and loss as FCOS (\cite{tian2019fcos}), replacing the backbone and feature pyramid network with five max pooling {layers}. {In our experiments, we also compare this tiny detector to a baseline: using three fully-connected layers to regress the ground-truth location taking the coordinates of high-confidence bounding boxes as input.}

\textbf{Loss of DRN.} After introducing our SCM, we can formulate the {full loss function for} DRN. We use $L_{rec}$, $L_{det}$, and $L_{scm}$ to indicate {the losses} from our recognition head, detection head, and spatial correlation module respectively. $\lambda_{rec}$, $\lambda_{det}$, and {$\lambda_{scm}$} are {the} regularization {hyperparameters} used to balance {the three separate loss functions}. We train our DRN using the following loss:
\begin{equation}
L = \lambda_{rec}L_{rec} + \lambda_{det}L_{det} + \lambda_{scm}L_{scm}
\label{eqn:total}
\end{equation}

\vspace{-0.05in}
\secvspace
\section{Experiments}
\secvspace
\subsecvspace
\subsection{PASCAL VOC}
\subsecvspace
\label{pascal}

\begin{table}[]
    \small
    \setlength{\tabcolsep}{1mm}{
    \begin{tabular}{c|c|cccccccccc>{\columncolor{mygray}}c}
    \hline
     & Base & \multicolumn{11}{c}{Novel}\tabularnewline
    Method & mean & table & dog & horse & mbike & person & plant & sheep & sofa & train & tv & \textbf{mean}\tabularnewline
    \hline \hline
    OICR & 42.1 & 33.4 & 29.3 & 56.3 & 64.6 & 8.0 & 23.5 & 47.2 & 47.2 & 48.3 & 61.7 & 42.0\tabularnewline
    PCL & 49.2 & 51.5 & 37.3 & 63.3 & 63.9 & 15.8 & 23.6 & 48.8 & 55.3 & 61.2 & 62.1 & 48.3\tabularnewline
    \hline 
    \hline
    MSD-VGG16 & 50.6 & 14.3 & 69.3 & 65.4 & 69.6 & 2.4 & 20.5 & 54.6 & 34.3 & 58.3 & 54.6 & 44.3\tabularnewline
    MSD-Ens & 53.4 & 18.3 & 70.6 & 66.7 & 69.8 & 3.7 & 24.7 & 55.0 & 37.4 & 58.3 & 57.3 & 46.1\tabularnewline
    MSD-Ens+FRCN & 53.9 & 15.3 & 72.0 & 74.4 & 65.2 & 15.4 & 25.1 & 53.6 & 54.4 & 45.6 & 61.4 & 48.2\tabularnewline
    Weight Transfer & 68.4 & 10.4 & 61.0 & 58.0 & 65.1 & \textbf{19.8} & 19.5 & 58.0 & 50.8 & 58.6 & 52.7 & 45.4\tabularnewline
    \hline 
    \hline 
    Finetune$^*$ & 71.8 & 17.8 & 22.9 & 15.2 & 71.2 & 10.2 & 15.1 & 61.7 & 36.6 & 21.9 & 61.3 & 33.4\tabularnewline
    Two Head$^*$ & \textbf{72.9} & 60.6 & 33.2 & 47.7 & 70.2 & 3.9 & 25.5 & 52.6 & 58.4 & 54.7 & 64.4 & 47.1\tabularnewline
    Two Head$^{+*}$ & 72.4 & 44.5 & 29.5 & 52.4 & 68.4 & 5.1 & 22.6 & 53.0 & 55.5 & 58.6 & 64.8 & 45.4\tabularnewline
    Two Branch$^*$ & 72.7 & 57.3 & 30.2 & 44.2 & 68.1 & 3.0 & 21.4 & 52.2 & 53.5& 51.2 & 59.7 & 44.1\tabularnewline
    \hline 
    \hline 
    Ours w/o SCM & 71.6 & 62.3 & 41.9 & 38.2 & \textbf{73.0} & 11.3 & 26.0 & 60.6 & \textbf{63.8} & \textbf{70.5} & 65.3 & 51.3\tabularnewline
    Ours & \textbf{72.9} & 61.0 & 57.1 & 63.5 & 72.0 & 19.5 & 24.2 & 60.9 & 58.6 & 68.5 & 65.5 & 55.1$\color{blue}{\mathbf{^{+3.8}}}$\tabularnewline
    Ours$^*$ w/o SCM & 72.7 & \textbf{66.8} & 50.4 & 57.0 & 71.5 & 12.1 & \textbf{27.6} & 57.1 & 62.7 & 54.2 & 64.2 & 52.4\tabularnewline
    Ours$^*$ & 72.7 & 60.9 & \textbf{59.4} & \textbf{70.5} & 71.0 & 17.5 & 24.1 & \textbf{62.0} & 60.5 & 62.4 & \textbf{69.1} & \textbf{55.7}$\color{blue}{\mathbf{^{+3.3}}}$\tabularnewline
    \end{tabular}
    }
    \caption{\textbf{Object Detection performance (mAP \%) on PASCAL VOC 2007 test set.} $^*$ indicates using the structure of OICR in the recognition head. "MSD-Ens" is the ensemble of AlexNet and VGG16. "MSD-Ens+FRCN" indicates using an ensemble model to predict pseudo ground truths and then {learn} a Fast-RCNN (\cite{girshick2015fast}) using VGG-16. }
    \label{tab:voc}
    \vspace{-0.2cm}
\end{table}

\textbf{Setup.} PASCAL VOC 2007 and 2012 datasets contain $9,962$ and $22,531$ images respectively for $20$ object classes. They are divided into train, val, and test sets. Here we follow previous work (\cite{oicr}) to choose the trainval set ($5,011$ images from 2007 and $11,540$ images from 2012). 
We divide the first 10 classes into base classes and the other 10 classes into novel classes.
To evaluate our methods, we calculate mean of Average Precision (mAP) based on the PASCAL criteria, \emph{i.e.}, IOU$>$0.5 between predicted boxes and ground truths. 

\textbf{Implementation details.} All {our} baselines, competitors and our framework are based on VGG16 (\cite{vgg}) followed most of weakly supervised object detection methods.  We set $\lambda_{rec}=1$, $\lambda_{det}=10$, and {$\lambda_{scm}=10$}. We train the whole framework for $20$ epochs using SGD with a momentum of $0.9$, a weight decay of $0.0005$ and a learning rate of $0.001$, which is {reduced by a factor} of 10 at $14^{th}$ epoch. {For a stable learning process}, {we don't provide} supervision from recognition head to detection head in the first $9$ epochs. 

\textbf{Baselines and competitors.} 
We compare against several baselines as mentioned in \cref{baseline}, two WSOD methods: OICR (\cite{oicr}) and PCL (\cite{tang2018pcl}), and two cross-supervised object detector: MSD (\cite{zhang2018mixed}), weight transfer (\cite{kuen2019scaling}). 

\textbf{Results.} As shown in Table \ref{tab:voc}, our method outperforms all other approaches by a large margin (over 7\% relative increase in mAP on novel classes). The results are consistent with our discussion in \cref{baseline}.
We note that (1) sharing backbone for the recognition and detection head learns a more discriminative embedding for novel objects.
In Table \ref{tab:voc}, Two Head$^*$ {boosts} the performance by 5 {points as} compared to only using the recognition head (OICR). (2) A supervision from recognition head to detection head exploits the full potential of a detection model. By adding the supervision (Ours$^*$ w/o SCM ), the result is improved by 5 {points} as compared to Two Head. (3) Our spatial correlation module successfully {captures} the spatial correlation between high-confidence proposals. It further {boosts} the performance by 3 {points}.

\subsecvspace
\subsection{COCO}
\subsecvspace
\label{coco}

\begin{table}[]
\small
\setlength{\tabcolsep}{0.9mm}{
\begin{tabular}{c|cccccc|cccccc}
\hline
                 & \multicolumn{6}{c|}{non-voc {$\rightarrow$} voc: test on B = \{voc\}} & \multicolumn{6}{c}{sixty {$\rightarrow$} twenty: test on B = \{twenty\}} \\
method           & AP      & AP$_{50}$    & AP$_{75}$   & AP$_S$   & AP$_M$   & AP$_L$   & AP    & AP$_{50}$   & AP$_{75}$   & AP$_S$   & AP$_M$   & AP$_L$   \\ \hline \hline
Rec. Head & 4.0     & 15.4         & 0.9         & 1.2      & 5.7      & 5.8      &   4.7    &    16.4         &      1.3       &    1.7      &   8.0       &  6.9        \\
OICR & 4.2     & 15.7         & 1.0         & 1.3      & 5.5      & 5.9      &   4.5    &    16.6         &      1.4       &    2.0      &   8.2       &  7.1 \\
PCL & 9.2     & 19.6         & -         & -      & -     & -     &   9.2    &  19.6         &    -         &    -      &    -      &     -     \\ \hline \hline

Weight T. & 9.3     & 26.4         & 5.7         & 5.8      & 11.7     & 12.4     &   8.7    &      25.5       &    5.5         &    5.4      &    11.5      &     11.7     \\ \hline \hline
Finetune         & 2.3     & 7.4          & 0.3         & 0.7      & 3.1      & 3.3      &   2.4    &    7.7         &     0.2        &     0.5     &   2.8       &     3.0     \\
Two Head         & 11.0    & 30.2         & 6.1         & 6.2      & 15.4     & 15.4     &  11.3     &     29.5        &    5.8         &   6.3       &   14.8       &     15.0     \\
Two Head$^+$     & 9.1     & 26.7         & 5.4         & 5.5      & 12.1     & 12.3     &   9.0    &     27.1        &      5.4       &    5.7      &   11.7       &   11.6       \\
Two Branch       & 9.4     & 26.6         & 5.6         & 5.7      & 12.3     & 12.4     &  8.5     &      24.4       &      4.5       &   4.3       &   11.9       &     11.9     \\ \hline \hline
\rowcolor{mygray}
Ours w/o SCM     & 12.5    & 33.6         & 6.6         & \textbf{7.3}      & \textbf{19.2}     & 16.4     &   12.6    &    32.3         &    7.8         &     7.0     &   \textbf{19.4}       &    17.4      \\
\rowcolor{mygray}
Ours             & \textbf{13.9}$\color{blue}{\mathbf{^{+1.4}}}$    & \textbf{36.2}$\color{blue}{\mathbf{^{+2.6}}}$         & \textbf{7.7}         & 6.9      & 18.8     & \textbf{19.9}     & \textbf{14.0}$\color{blue}{\mathbf{^{+1.4}}}$     &     \textbf{34.5}$\color{blue}{\mathbf{^{+2.2}}}$        &   \textbf{8.9}      &    \textbf{7.1}      &    19.2      &     \textbf{20.6}    
\end{tabular}
}
\caption{\label{tab:coco} \textbf{The results on COCO.} We compare our method with several strong baselines in \cref{baseline} and competitors. Our method significantly outperforms these approaches, showing that our cross-supervised object detector is capable of detecting novel objects in complex multi-object scenes.}

\vspace{-0.2cm}
\end{table}

\begin{table}[!b]
    \centering
    \small
    \setlength{\tabcolsep}{0.9mm}{
    \begin{subtable}[t]{0.47\linewidth}
        \centering
        \small    
        \begin{tabular}{c|c|c}
                    & non-voc{$\rightarrow$}voc & sixty{$\rightarrow$}twenty \\
        \multicolumn{1}{c|}{\ \ \ \ \ \ \ \ \ method\ \ \ \ \ \ \ \ \ }   & AP$_{50}$ on B         & AP$_{50}$ on B     \\ \hline
        {\texttt{max}}         & 35.5                      &           33.8                \\
        {\texttt{sum}}         & 36.0                      &           34.0                \\
        {\texttt{num}}         & 31.5                      &            29.5               \\ \hline
        \rowcolor{mygray}
        {\texttt{max+sum}}     & \textbf{36.2}                      &     \textbf{34.5}                      \\
        {\texttt{max+num}}     & 35.7                      &         34.1                  \\ 
        {\texttt{sum+num}}     & 35.9                      &          34.2                 \\
        \hline
        {\texttt{max+sum+num}} & 36.1                      &          34.2                
        \end{tabular}
        \caption{\textbf{Ablation on {Heatmap synthesis}.} The result suggests using two-channel heatmap consists of maximum confidence and sum of confidence {over proposals covering that position.}}
        \label{tab:a}
    \end{subtable}%
    \hspace*{1.5em}
    \begin{subtable}[t]{0.47\linewidth}
        \centering
        \small
        \begin{tabular}[h]{cccc}
                                  & \multicolumn{1}{c|}{}         & \multicolumn{1}{c|}{non-voc{$\rightarrow$}voc} & sixty{$\rightarrow$}twenty \\
        \multicolumn{2}{c|}{method}                               & \multicolumn{1}{c|}{AP$_{50}$ on B}         & AP$_{50}$ on B     \\ \hline
        \multirow{3}{*}{Fc layer} & \multicolumn{1}{c|}{2 layer}  & \multicolumn{1}{c|}{31.0}                      &       28.7                    \\
                                  & \multicolumn{1}{c|}{3 layer}  & \multicolumn{1}{c|}{30.8}                      & \multicolumn{1}{c}{28.3}      \\
                                  & \multicolumn{1}{c|}{4 layer}  & \multicolumn{1}{c|}{30.5}                      &               28.5            \\ \hline
             & \multicolumn{1}{c|}{R-50-FPN} & \multicolumn{1}{c|}{\textbf{36.4}}                      &    \textbf{34.8}                       \\
             & \multicolumn{1}{c|}{4 conv}   & \multicolumn{1}{c|}{35.8}                      &         33.8                  \\ 
        \rowcolor{mygray}
        \multirow{-3}{*}{FCOS}
                                  & \multicolumn{1}{c|}{5 conv}   & \multicolumn{1}{c|}{36.2}                      &         34.5                  \\ \hline
        \multicolumn{2}{c|}{w/o SCM}                              & \multicolumn{1}{c|}{33.6}                      &          32.3          
        \end{tabular}
        \caption{\textbf{Ablation on the structure of SCM.} FCOS with 5 conv {layers} {has nearly} the best performance {and very} few parameters {compared} to a ResNet-50 backbone.}
        \label{tab:b}
    \end{subtable}%
    
    \begin{subtable}[t]{0.47\linewidth}
        \centering
        \small    
        \begin{tabular}{ccc}
        \multicolumn{1}{c|}{}       & \multicolumn{1}{c|}{non-voc{$\rightarrow$}voc} & sixty{$\rightarrow$}twenty \\
        \multicolumn{1}{c|}{\ \ \ \ \ \ \ \ \ method\ \ \ \ \ \ \ \ \ } & \multicolumn{1}{c|}{AP$_{50}$ on B}         & AP$_{50}$ on B     \\ \hline
        \multicolumn{1}{c|}{WSDDN}  & \multicolumn{1}{c|}{35.7}                      &      33.8                     \\
        \multicolumn{1}{c|}{OICR}   & \multicolumn{1}{c|}{\textbf{36.6}}                      &  \textbf{34.7}                         \\ \hline
        \rowcolor{mygray}
        \multicolumn{1}{c|}{Ours}   & \multicolumn{1}{c|}{36.4}                      &   34.5  \\
        \multicolumn{1}{c}{}   & \multicolumn{1}{c}{}              &  
        \end{tabular}
        \caption{\textbf{Ablation on the structure of the recognition head.} OICR {has} more {refinement} branches so it behaves a little better than our recognition head but {takes} double {the} computation time.}
        \label{tab:c}
    \end{subtable}%
    \hspace*{1.5em}
    \begin{subtable}[t]{0.47\linewidth}
        \centering
        \small 
        \begin{tabular}{cc|cc}
                                            &  & \multicolumn{2}{c}{base{$\rightarrow$}novel}\\
        dataset                                    & method & AP$_{50}$ on A & AP$_{50}$ on B\\ \hline
                        & RPN    & \textbf{76.2} & 46.1 \\
        \rowcolor{mygray}
        \multirow{-2}{*}{PASCAL VOC}             & SS     & 72.7 & \textbf{55.7} \\ \hline
        \rowcolor{mygray}
        \multirow{2}{*}{non-voc{$\rightarrow$}voc} & RPN    & \textbf{46.3} & \textbf{36.2} \\
                                                  & SS     & 42.5 & 34.5 \\    
        \end{tabular}
        \caption{\textbf{Ablation on the proposal generator.} On PASCAL VOC, there {are not} enough categories to learn a good RPN. So, we use selective search and RPN to generate proposals for PASCAL VOC and COCO respectively.}
        \label{tab:d}
    \end{subtable}%
    }
    
    \label{tab:ablation}
    \caption{\textbf{Ablation study of our method.}}
\end{table}

\textbf{Setup.} We train on the COCO \emph{train2017} split and test on \emph{val2017} split. We simulate the cross-supervised object detection scenario on COCO by splitting the 80 classes into base and novel classes. We use a 20/60 split same as \cite{hu2018learning}, dividing the COCO categories into {all the 20 classes} contained in PASCAL VOC and the 60 that are not. We refer to these as the `voc' and `non-voc' category sets. `voc$\rightarrow$non-voc' indicates that we take `voc' as our base classes and `non-voc' as our novel classes. Similarly, we split the first 20 classes into `twenty' and the last 60 classes into 'sixty'.

\textbf{Implementation details.} The implementation details are the same as \cref{pascal} {by} default. We train the whole framework for $13$ epochs. There is no supervision from recognition head to detection head in the first $5$ epochs. The learning rate is {reduced by a factor of} 10 at $8^{th}$, and $12^{th}$ {epochs}.

\textbf{{Baselines} and competitors.} Most baselines and competitors are the same as \cref{pascal}. {'Rec. Head'} represents only using our recognition head structure as a weakly supervised object detector.  

\textbf{Results.} The results on COCO still support our discussion in \cref{pascal}. Even in complex multi objects scenes, {our} DRN outperforms all baselines and competitors by a large margin. 

\subsecvspace
\subsection{Ablation Experiments}
\subsecvspace
\label{ablation}

\begin{figure*}
  \centering
  \includegraphics[width=13.5cm]{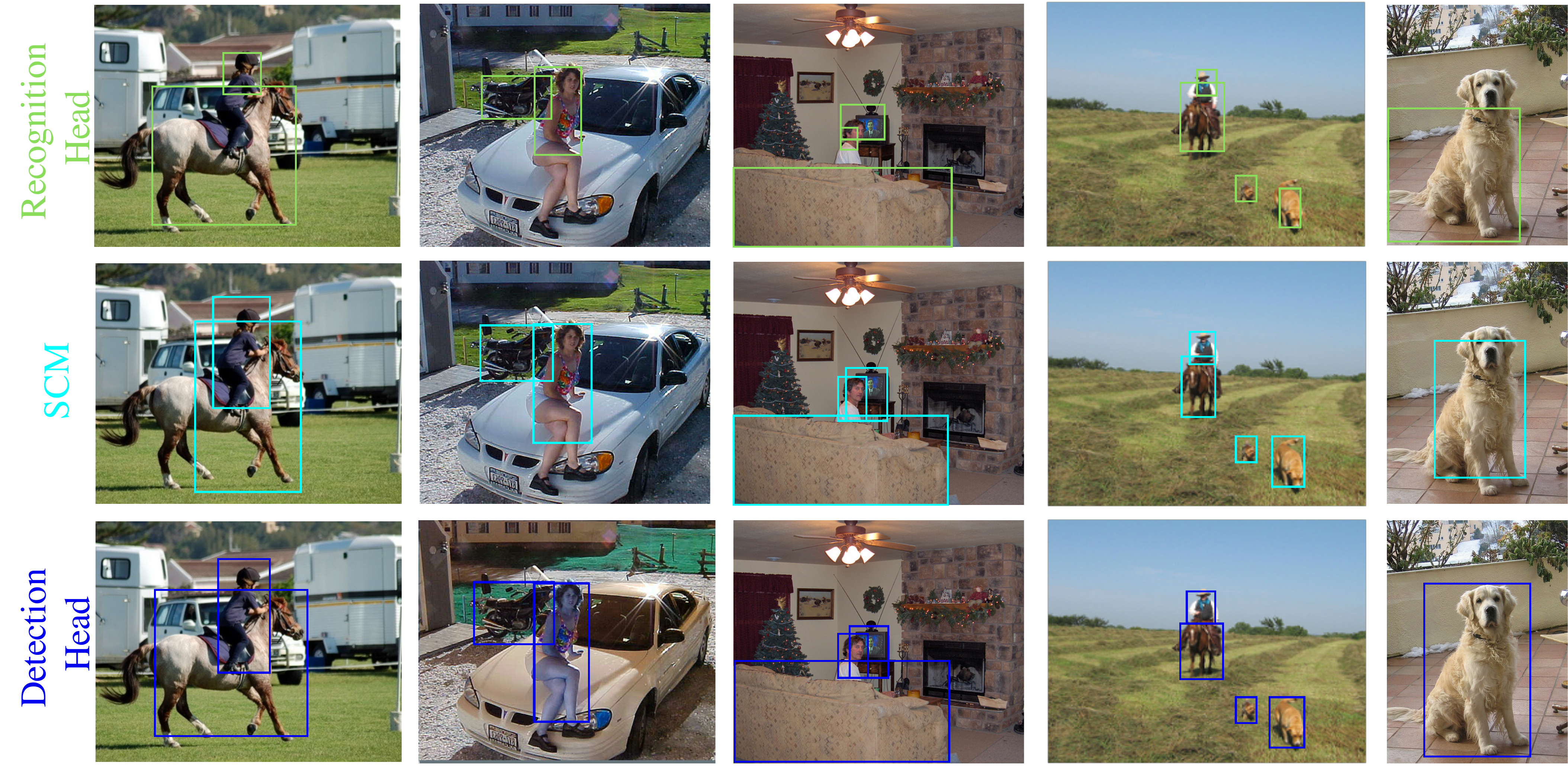}
  \caption{\textbf{Detection results on \emph{novel} objects.} The results are from our proposed model but with different heads. The first row shows the {results} of the recognition head. The second row lists the {results} from SCM. The third row {displays} the {results} from the detection head.}
  \label{fig:show}
  \vspace{-0.2cm}
\end{figure*}

{\textbf{Heatmap synthesis.}} In Table ~\ref{tab:a}, we {compare the different methods to synthesize the heatmaps in} the spatial correlation module.
For each position in the heatmap, {we consider three kinds of values:} the \texttt{max}imum of confidence, the \texttt{sum} of confidence, and the \texttt{num}ber of proposals covering the position. This result informs us to use {\texttt{max}} and {\texttt{sum}} to create a two-channel heatmap. 


\textbf{Structure of SCM.} In Table \ref{tab:b}, we compare different implementations of SCM. We compare the FCOS (\cite{tian2019fcos}) with 5 convolutional layers and the standard FCOS with a ResNet-50 (\cite{he2016deep}) backbone. We also compare to the regression baseline mentioned in \cref{SCM}. Considering the computation cost, we choose FCOS with 5 convolutional layer as our heatmap detector.

\textbf{Structure of the Recognition head.} In Table \ref{tab:c}, we compare different structures {for} the recognition head. WSDDN (\cite{bilen2016wsddn}) and OICR are compared to our structure. The results support that our model can benefit from a stronger recognition head.  

\textbf{Different proposal generation methods.} Table \ref{tab:d} shows the ablation of different ways to generate proposals. 
In PASCAL VOC with only 10 base classes, RPN performs worse than selective search. In COCO with 60 base classes, RPN performs better than selective search. 

\textbf{Visualization.}  Fig.~\ref{fig:show} shows detection results on novel objects. Images in the first row, the second row, and the third row are detected by our model from the recognition head, the SCM, and the detection head respectively. The images in the first row tend to focus on the discriminating {parts} of the {objects}, e.g. the first and {the} second {images} {contain only} a part of the person. It also tends to detect co-occurring objects, e.g. the fourth {image} not only {detects} horse but also a large part of the person. Our SCM {alleviates} these problems. It tends to focus on the whole object, e.g. the first and the third samples detect the whole person instead of only {the head}. Also, it can correct unsatisfactory bounding boxes {distracted} by co-occurring objects, e.g. SCM correctly {localizes the} horse instead of {localizing} both {the} person and {the} horse in the fourth example. Obviously, bounding boxes in the third row are the best, indicating the efficacy of our framework.

\secvspace
\section{Conclusion}
\secvspace

We focus on a novel learning paradigm---cross-supervised object detection. We explore two major ways to build a good cross-supervised object detector: sharing network backbone between a recognition head and a detection head, and learning a spatial correlation module to bridge the gap between recognition and detection. Significant improvement on PASCAL VOC and COCO suggests a novel and promising approach for expanding object detection to a much larger number of categories.

\bibliography{iclr2020_conference}

\begin{thebibliography}{37}
\providecommand{\natexlab}[1]{#1}
\providecommand{\url}[1]{\texttt{#1}}
\expandafter\ifx\csname urlstyle\endcsname\relax
  \providecommand{\doi}[1]{doi: #1}\else
  \providecommand{\doi}{doi: \begingroup \urlstyle{rm}\Url}\fi

\bibitem[Arun et~al.(2019)Arun, Jawahar, and Kumar]{Arun_2019_CVPR}
Aditya Arun, C.V. Jawahar, and M.~Pawan Kumar.
\newblock Dissimilarity coefficient based weakly supervised object detection.
\newblock In \emph{The IEEE Conference on Computer Vision and Pattern
  Recognition (CVPR)}, June 2019.

\bibitem[Bilen \& Vedaldi(2016)Bilen and Vedaldi]{bilen2016wsddn}
Hakan Bilen and Andrea Vedaldi.
\newblock Weakly supervised deep detection networks.
\newblock In \emph{Proceedings of the IEEE Conference on Computer Vision and
  Pattern Recognition}, pp.\  2846--2854, 2016.

\bibitem[Diba et~al.(2017)Diba, Sharma, Pazandeh, Pirsiavash, and
  Van~Gool]{diba2017weakly}
Ali Diba, Vivek Sharma, Ali Pazandeh, Hamed Pirsiavash, and Luc Van~Gool.
\newblock Weakly supervised cascaded convolutional networks.
\newblock In \emph{Proceedings of the IEEE conference on computer vision and
  pattern recognition}, pp.\  914--922, 2017.

\bibitem[Dietterich et~al.(1997)Dietterich, Lathrop, and
  Lozano-P{\'e}rez]{dietterich1997solving}
Thomas~G Dietterich, Richard~H Lathrop, and Tom{\'a}s Lozano-P{\'e}rez.
\newblock Solving the multiple instance problem with axis-parallel rectangles.
\newblock \emph{Artificial intelligence}, 89\penalty0 (1-2):\penalty0 31--71,
  1997.

\bibitem[Everingham et~al.(2010)Everingham, Van~Gool, Williams, Winn, and
  Zisserman]{everingham2010pascal}
Mark Everingham, Luc Van~Gool, Christopher~KI Williams, John Winn, and Andrew
  Zisserman.
\newblock The pascal visual object classes (voc) challenge.
\newblock \emph{International journal of computer vision}, 88\penalty0
  (2):\penalty0 303--338, 2010.

\bibitem[Gao et~al.(2019{\natexlab{a}})Gao, Wang, Dai, Li, and
  Nevatia]{gao2019note}
Jiyang Gao, Jiang Wang, Shengyang Dai, Li-Jia Li, and Ram Nevatia.
\newblock Note-rcnn: Noise tolerant ensemble rcnn for semi-supervised object
  detection.
\newblock In \emph{Proceedings of the IEEE International Conference on Computer
  Vision}, pp.\  9508--9517, 2019{\natexlab{a}}.

\bibitem[Gao et~al.(2019{\natexlab{b}})Gao, Liu, Guo, Ye, Wan, You, and
  Fan]{gao2019c}
Yan Gao, Boxiao Liu, Nan Guo, Xiaochun Ye, Fang Wan, Haihang You, and Dongrui
  Fan.
\newblock C-midn: Coupled multiple instance detection network with segmentation
  guidance for weakly supervised object detection.
\newblock In \emph{Proceedings of the IEEE International Conference on Computer
  Vision}, pp.\  9834--9843, 2019{\natexlab{b}}.

\bibitem[Girshick(2015)]{girshick2015fast}
Ross Girshick.
\newblock Fast r-cnn.
\newblock In \emph{Proceedings of the IEEE international conference on computer
  vision}, pp.\  1440--1448, 2015.

\bibitem[He et~al.(2016)He, Zhang, Ren, and Sun]{he2016deep}
Kaiming He, Xiangyu Zhang, Shaoqing Ren, and Jian Sun.
\newblock Deep residual learning for image recognition.
\newblock In \emph{Proceedings of the IEEE conference on computer vision and
  pattern recognition}, pp.\  770--778, 2016.

\bibitem[Hoffman et~al.(2014)Hoffman, Guadarrama, Tzeng, Hu, Donahue, Girshick,
  Darrell, and Saenko]{hoffman2014lsda}
Judy Hoffman, Sergio Guadarrama, Eric~S Tzeng, Ronghang Hu, Jeff Donahue, Ross
  Girshick, Trevor Darrell, and Kate Saenko.
\newblock Lsda: Large scale detection through adaptation.
\newblock In \emph{Advances in Neural Information Processing Systems}, pp.\
  3536--3544, 2014.

\bibitem[Hoffman et~al.(2015)Hoffman, Pathak, Darrell, and
  Saenko]{hoffman2015detector}
Judy Hoffman, Deepak Pathak, Trevor Darrell, and Kate Saenko.
\newblock Detector discovery in the wild: Joint multiple instance and
  representation learning.
\newblock In \emph{Proceedings of the ieee conference on computer vision and
  pattern recognition}, pp.\  2883--2891, 2015.

\bibitem[Hu et~al.(2018)Hu, Doll{\'a}r, He, Darrell, and
  Girshick]{hu2018learning}
Ronghang Hu, Piotr Doll{\'a}r, Kaiming He, Trevor Darrell, and Ross Girshick.
\newblock Learning to segment every thing.
\newblock In \emph{Proceedings of the IEEE Conference on Computer Vision and
  Pattern Recognition}, pp.\  4233--4241, 2018.

\bibitem[Kosugi et~al.(2019)Kosugi, Yamasaki, and Aizawa]{kosugi2019object}
Satoshi Kosugi, Toshihiko Yamasaki, and Kiyoharu Aizawa.
\newblock Object-aware instance labeling for weakly supervised object
  detection.
\newblock In \emph{Proceedings of the IEEE International Conference on Computer
  Vision}, pp.\  6064--6072, 2019.

\bibitem[Kuen et~al.(2019)Kuen, Perazzi, Lin, Zhang, and Tan]{kuen2019scaling}
Jason Kuen, Federico Perazzi, Zhe Lin, Jianming Zhang, and Yap-Peng Tan.
\newblock Scaling object detection by transferring classification weights.
\newblock In \emph{Proceedings of the IEEE International Conference on Computer
  Vision}, pp.\  6044--6053, 2019.

\bibitem[Li et~al.(2019)Li, Kan, Shan, and Chen]{li2019weakly}
Xiaoyan Li, Meina Kan, Shiguang Shan, and Xilin Chen.
\newblock Weakly supervised object detection with segmentation collaboration.
\newblock In \emph{Proceedings of the IEEE International Conference on Computer
  Vision}, pp.\  9735--9744, 2019.

\bibitem[Lin et~al.(2014)Lin, Maire, Belongie, Hays, Perona, Ramanan,
  Doll{\'a}r, and Zitnick]{COCO}
Tsung-Yi Lin, Michael Maire, Serge Belongie, James Hays, Pietro Perona, Deva
  Ramanan, Piotr Doll{\'a}r, and C~Lawrence Zitnick.
\newblock Microsoft coco: Common objects in context.
\newblock In \emph{European conference on computer vision}, pp.\  740--755.
  Springer, 2014.

\bibitem[Lin et~al.(2017)Lin, Goyal, Girshick, He, and Doll{\'a}r]{focalLoss}
Tsung-Yi Lin, Priya Goyal, Ross Girshick, Kaiming He, and Piotr Doll{\'a}r.
\newblock Focal loss for dense object detection.
\newblock In \emph{Proceedings of the IEEE international conference on computer
  vision}, pp.\  2980--2988, 2017.

\bibitem[Redmon et~al.(2016)Redmon, Divvala, Girshick, and Farhadi]{YOLO}
Joseph Redmon, Santosh Divvala, Ross Girshick, and Ali Farhadi.
\newblock You only look once: Unified, real-time object detection.
\newblock In \emph{Proceedings of the IEEE conference on computer vision and
  pattern recognition}, pp.\  779--788, 2016.

\bibitem[Ren et~al.(2015)Ren, He, Girshick, and Sun]{faster_rcnn}
Shaoqing Ren, Kaiming He, Ross Girshick, and Jian Sun.
\newblock Faster r-cnn: Towards real-time object detection with region proposal
  networks.
\newblock In \emph{Advances in neural information processing systems}, pp.\
  91--99, 2015.

\bibitem[Russakovsky et~al.(2015)Russakovsky, Deng, Su, Krause, Satheesh, Ma,
  Huang, Karpathy, Khosla, Bernstein, Berg, and Fei-Fei]{ILSVRC15}
Olga Russakovsky, Jia Deng, Hao Su, Jonathan Krause, Sanjeev Satheesh, Sean Ma,
  Zhiheng Huang, Andrej Karpathy, Aditya Khosla, Michael Bernstein,
  Alexander~C. Berg, and Li~Fei-Fei.
\newblock {ImageNet Large Scale Visual Recognition Challenge}.
\newblock \emph{International Journal of Computer Vision (IJCV)}, 115\penalty0
  (3):\penalty0 211--252, 2015.
\newblock \doi{10.1007/s11263-015-0816-y}.

\bibitem[Simonyan \& Zisserman(2015)Simonyan and Zisserman]{vgg}
Karen Simonyan and Andrew Zisserman.
\newblock Very deep convolutional networks for large-scale image recognition.
\newblock In \emph{International Conference on Learning Representations}, 2015.

\bibitem[Tang et~al.(2017)Tang, Wang, Bai, and Liu]{oicr}
Peng Tang, Xinggang Wang, Xiang Bai, and Wenyu Liu.
\newblock Multiple instance detection network with online instance classifier
  refinement.
\newblock In \emph{Proceedings of the IEEE Conference on Computer Vision and
  Pattern Recognition}, pp.\  2843--2851, 2017.

\bibitem[Tang et~al.(2018)Tang, Wang, Bai, Shen, Bai, Liu, and
  Yuille]{tang2018pcl}
Peng Tang, Xinggang Wang, Song Bai, Wei Shen, Xiang Bai, Wenyu Liu, and Alan
  Yuille.
\newblock Pcl: Proposal cluster learning for weakly supervised object
  detection.
\newblock \emph{IEEE transactions on pattern analysis and machine
  intelligence}, 42\penalty0 (1):\penalty0 176--191, 2018.

\bibitem[Tang et~al.(2016)Tang, Wang, Gao, Dellandr{\'e}a, Gaizauskas, and
  Chen]{tang2016large}
Yuxing Tang, Josiah Wang, Boyang Gao, Emmanuel Dellandr{\'e}a, Robert
  Gaizauskas, and Liming Chen.
\newblock Large scale semi-supervised object detection using visual and
  semantic knowledge transfer.
\newblock In \emph{Proceedings of the IEEE Conference on Computer Vision and
  Pattern Recognition}, pp.\  2119--2128, 2016.

\bibitem[Tian et~al.(2019)Tian, Shen, Chen, and He]{tian2019fcos}
Zhi Tian, Chunhua Shen, Hao Chen, and Tong He.
\newblock Fcos: Fully convolutional one-stage object detection.
\newblock In \emph{Proceedings of the IEEE International Conference on Computer
  Vision}, pp.\  9627--9636, 2019.

\bibitem[Uijlings et~al.(2018)Uijlings, Popov, and
  Ferrari]{uijlings2018revisiting}
Jasper Uijlings, Stefan Popov, and Vittorio Ferrari.
\newblock Revisiting knowledge transfer for training object class detectors.
\newblock In \emph{Proceedings of the IEEE Conference on Computer Vision and
  Pattern Recognition}, pp.\  1101--1110, 2018.

\bibitem[Uijlings et~al.(2013)Uijlings, Van De~Sande, Gevers, and
  Smeulders]{uijlings2013selective}
Jasper~RR Uijlings, Koen~EA Van De~Sande, Theo Gevers, and Arnold~WM Smeulders.
\newblock Selective search for object recognition.
\newblock \emph{International journal of computer vision}, 104\penalty0
  (2):\penalty0 154--171, 2013.

\bibitem[Wan et~al.(2018)Wan, Wei, Jiao, Han, and Ye]{wan2018min}
Fang Wan, Pengxu Wei, Jianbin Jiao, Zhenjun Han, and Qixiang Ye.
\newblock Min-entropy latent model for weakly supervised object detection.
\newblock In \emph{Proceedings of the IEEE Conference on Computer Vision and
  Pattern Recognition}, pp.\  1297--1306, 2018.

\bibitem[Wan et~al.(2019)Wan, Liu, Ke, Ji, Jiao, and Ye]{CMIL}
Fang Wan, Chang Liu, Wei Ke, Xiangyang Ji, Jianbin Jiao, and Qixiang Ye.
\newblock C-mil: Continuation multiple instance learning for weakly supervised
  object detection.
\newblock In \emph{The IEEE Conference on Computer Vision and Pattern
  Recognition (CVPR)}, June 2019.

\bibitem[Wei et~al.(2018)Wei, Shen, Cheng, Shi, Xiong, Feng, and
  Huang]{wei2018ts2c}
Yunchao Wei, Zhiqiang Shen, Bowen Cheng, Honghui Shi, Jinjun Xiong, Jiashi
  Feng, and Thomas Huang.
\newblock Ts2c: Tight box mining with surrounding segmentation context for
  weakly supervised object detection.
\newblock In \emph{Proceedings of the European Conference on Computer Vision
  (ECCV)}, pp.\  434--450, 2018.

\bibitem[Yang et~al.(2019)Yang, Li, and Dou]{yang2019towards}
Ke~Yang, Dongsheng Li, and Yong Dou.
\newblock Towards precise end-to-end weakly supervised object detection
  network.
\newblock In \emph{Proceedings of the IEEE International Conference on Computer
  Vision}, pp.\  8372--8381, 2019.

\bibitem[Zeng et~al.(2019)Zeng, Liu, Fu, Chao, and Zhang]{zeng2019wsod2}
Zhaoyang Zeng, Bei Liu, Jianlong Fu, Hongyang Chao, and Lei Zhang.
\newblock Wsod2: Learning bottom-up and top-down objectness distillation for
  weakly-supervised object detection.
\newblock In \emph{Proceedings of the IEEE International Conference on Computer
  Vision}, pp.\  8292--8300, 2019.

\bibitem[Zhang et~al.(2018{\natexlab{a}})Zhang, Huang, Zhang,
  et~al.]{zhang2018mixed}
Junge Zhang, Kaiqi Huang, Jianguo Zhang, et~al.
\newblock Mixed supervised object detection with robust objectness transfer.
\newblock \emph{IEEE transactions on pattern analysis and machine
  intelligence}, 41\penalty0 (3):\penalty0 639--653, 2018{\natexlab{a}}.

\bibitem[Zhang et~al.(2018{\natexlab{b}})Zhang, Feng, Xiong, and
  Tian]{zhang2018zigzag}
Xiaopeng Zhang, Jiashi Feng, Hongkai Xiong, and Qi~Tian.
\newblock Zigzag learning for weakly supervised object detection.
\newblock In \emph{Proceedings of the IEEE Conference on Computer Vision and
  Pattern Recognition}, pp.\  4262--4270, 2018{\natexlab{b}}.

\bibitem[Zhang et~al.(2018{\natexlab{c}})Zhang, Bai, Ding, Li, and
  Ghanem]{zhang2018w2f}
Yongqiang Zhang, Yancheng Bai, Mingli Ding, Yongqiang Li, and Bernard Ghanem.
\newblock W2f: A weakly-supervised to fully-supervised framework for object
  detection.
\newblock In \emph{Proceedings of the IEEE Conference on Computer Vision and
  Pattern Recognition}, pp.\  928--936, 2018{\natexlab{c}}.

\bibitem[Zhou et~al.(2016)Zhou, Khosla, Lapedriza, Oliva, and Torralba]{CAM}
Bolei Zhou, Aditya Khosla, Agata Lapedriza, Aude Oliva, and Antonio Torralba.
\newblock Learning deep features for discriminative localization.
\newblock In \emph{Proceedings of the IEEE conference on computer vision and
  pattern recognition}, pp.\  2921--2929, 2016.

\bibitem[Zhu et~al.(2017)Zhu, Zhou, Ye, Qiu, and Jiao]{zhu2017soft}
Yi~Zhu, Yanzhao Zhou, Qixiang Ye, Qiang Qiu, and Jianbin Jiao.
\newblock Soft proposal networks for weakly supervised object localization.
\newblock In \emph{Proceedings of the IEEE International Conference on Computer
  Vision}, pp.\  1841--1850, 2017.

\end{thebibliography}
\bibliographystyle{iclr2020_conference}


\end{document}